\documentclass[11pt]{article}

\usepackage[final]{acl}

\usepackage{times}
\usepackage{latexsym}

\usepackage[T1]{fontenc}

\usepackage[utf8]{inputenc}

\usepackage{microtype}

\usepackage{inconsolata}

\usepackage{graphicx}

\usepackage{makecell}
\usepackage{amsmath}
\usepackage{booktabs}

\title{Why Are We Lonely? Leveraging LLMs to Measure and Understand Loneliness in Caregivers and Non-caregivers}

\author{
  \textbf{Michelle Damin Kim\textsuperscript{1}},
  \textbf{Ellie S. Paek\textsuperscript{1}},
  \textbf{Yufen Lin\textsuperscript{2}},
  \textbf{Emily Mroz\textsuperscript{2}},
  \textbf{Jane Chung\textsuperscript{2}},
\\
  \textbf{Jinho D. Choi\textsuperscript{1}},
\\
  \textsuperscript{1}Department of Computer Science, Emory University, Atlanta, GA, USA,\\
  \textsuperscript{2}Nell Hodgson Woodruff School of Nursing, Emory University, Atlanta, GA, USA
\\
  \texttt{\{michelle.kim2, ellie.paek, yufen.lin, emily.mroz, jane.chung}
  \\
  \texttt{ jinho.choi\}@emory.edu} 
}

\begin{document}
\maketitle
\begin{abstract}
This paper presents an LLM-driven approach for constructing diverse social media datasets to measure and compare loneliness in the caregiver and non-caregiver populations. We introduce an expert-developed loneliness evaluation framework and an expert-informed typology for categorizing causes of loneliness for analyzing social media text. Using a human-validated data processing pipeline, we apply GPT-4o, GPT-5-nano, and GPT-5 to build a high-quality Reddit corpus and analyze loneliness across both populations. The loneliness evaluation framework achieved average accuracies of 76.09\% and 79.78\% for caregivers and non-caregivers, respectively. The cause categorization framework achieved micro-aggregate F1 scores of 0.825 and 0.80 for caregivers and non-caregivers, respectively. Across populations, we observe substantial differences in the distribution of types of causes of loneliness. Caregivers’ loneliness were predominantly linked to caregiving roles, identity recognition, and feelings of abandonment, indicating distinct loneliness experiences between the two groups. Demographic extraction further demonstrates the viability of Reddit for building a diverse caregiver loneliness dataset. Overall, this work establishes an LLM-based pipeline for creating high quality social media datasets for studying loneliness and demonstrates its effectiveness in analyzing population-level differences in the manifestation of loneliness.
\end{abstract}

\section{Introduction}
Caregivers, people who provide ongoing care to another as a result of a medical condition, are a growing population in the United States. From 43.5 million estimated caregivers in 2015 to 63 million in 2025, 1 in 4 American adults are caregivers \citep{national-alliance-2025}. Caregivers face a heightened risk of loneliness. Loneliness, a public health concern, is associated with health challenges such as depression, sleep disorders, and increased risk of obesity and diabetes \citep{bonin-2021, gray-2019}. While caregivers are a significant and increasing population, there remains a lack of specialized frameworks for understanding loneliness in caregivers \citep{mcrae-2009, victor-2021, vasileiou-2017}.

Following the advent of Large Language Models (LLMs), there is still a dearth of quantitative investigation that applies state-of-the-art models to uncover new typologies, understandings, and facets of caregiver loneliness across contexts. While datasets for loneliness detection and categorization are available, there are limited available datasets for the study of loneliness in caregivers that enable the application of psychologically grounded frameworks designed for a social media and caregiving context \citep{yang-2023, jiang-2022, fujikawa-2024}.

Thus, this work leverages LLMs and expert-informed frameworks to analyze loneliness social media data, enabling comparison of the manifestation of loneliness across the caregiver and non-caregiver populations.

We are guided by the following research questions:
\begin{enumerate}
\item Is it feasible to utilize Reddit posts to study loneliness, particularly in caregivers, by building a high quality and diverse dataset?

\item Are LLMs capable of being applied to effectively analyze loneliness in accordance with psychologically grounded and expert-informed guidelines? 

\item Through the application of LLMs, can we compare the experience of loneliness across the caregiver and non-caregiver populations?
\end{enumerate}
This work produced the following key contributions:
\begin{enumerate}
\item A validated data processing pipeline for evaluating and analyzing loneliness in Reddit data.

\item A high quality dataset of Reddit posts from the caregiver and non-caregiver populations, with extracted measurements of loneliness, categorized cause(s) of loneliness, and demographic information. 

\item An expert-produced adaptation of the UCLA Loneliness scale for measuring loneliness in social media data. 

\item An expert-informed framework for categorizing causes of loneliness that reflects themes of caregiver loneliness.
\end{enumerate}

\section{Related Work}

While NLP has been applied to a variety of mental health contexts, application to the assessment or analysis of loneliness has been limited \citep{yang-2023}. One challenge explored by prior works is handling ambiguity in both defining and categorizing loneliness via a psychologically grounded framework. A frequently cited definition of loneliness is that loneliness is a situation where there is an unpleasant lack of or quality of relationships, including when one's desired amount of relationships or level of intimacy in relationships is less than one's actuality \citep{gierveld-1998}. Other definitions add an aspect of perceived inability to attain one's desired state of connection, emphasizing the importance of internal subjective experience in the experience of loneliness \citep{motta-2021}.  

Existing works have worked toward building loneliness datasets, detecting loneliness, and analyzing loneliness. \citet{fujikawa-2024} worked toward creating a Japanese language loneliness dataset with annotation guidelines for the intensity of loneliness, using a pre-trained BERT model to classify whether text expressed loneliness or not and whether the intensity was strong or weak. 

For measuring loneliness, the UCLA Loneliness Scale and the De Jong Gierveld Loneliness Scale are frequently used \citep{gierveld-1985, gierveld-2006}. The UCLA Loneliness Scale introduces a 20-item scale where each item is a first person statement regarding how an individual feels, relating to loneliness, and is rated based on one's frequency of feeling like the statement \citep{russell-1978}. \citet{garg-2023} applied multiple loneliness scales and explored the appropriate amount of inference 
to annotate Reddit data for detecting loneliness and the cause of the loneliness, resulting in the LonXplain dataset. This work tested the performance of Word2Vec, GloVe, and Gated Recurrent Units for a binary loneliness detection task with GloVe and GRU achieving an F1 score 0.77 and an accuracy of 0.78. Leveraging Local Interpretable Model-Agnostic Explanations (LIME), this work demonstrates the interest using textual evidence to categorize loneliness.

When analyzing and categorizing types of loneliness, there is not a comprehensive, widely accepted, detailed framework. Two commonly used types are "social," loneliness from a lacking quantity of relationships, and "emotional," loneliness from a lack of close relationships \citep{russell-1984}. Other categorizations include characterizing loneliness based on temporality, distinguishing transient lonely moods, chronic loneliness, and situational loneliness \citep{motta-2021}. \citet{jiang-2022} applied 2 BERT-based models for binary loneliness classification and fine-grained category classification to Reddit posts. This work compared loneliness expression across 2 settings: young adult and loneliness related subreddits. They used a schema with four main categories: duration, loneliness context, mentioned interpersonal relationships, and coping strategy type. This work proved the potential of automatic detection of loneliness categorization and demonstrated a model framework for capturing multidimensional aspects of loneliness. 

However, there is limited exploration of building high quality datasets that enable in-depth analysis of loneliness, particularly in the LLM-based application of expert-created loneliness evaluation and categorization frameworks. Within the caregiver population, experiences of loneliness are strongly influenced by the caregiving role. Four themes of loneliness that arise from caregiving include: (1) a loss of psychological, personal, and physical space due to caregiving responsibilities (2) loss of and changes to close relationships, such as with the care recipient (3) a lack of social recognition of the caregiver role and difficulty relating with non-caregivers (4) feelings of helplessness to improve their situation due to a sense of sole responsibility \citep{vasileiou-2017}. Thus, frameworks that recognize population-specific aspects of loneliness are needed to build and analyze population-specific datasets. Our work builds on these prior works to detect, measure, and analyze caregiver and non-caregiver loneliness from Reddit posts to investigate the application of tailored and complex annotation frameworks via a validated LLM-powered data processing pipeline. 

\section{Methodology}
\subsection{Data Collection}

We used the Python Reddit API Wrapper (PRAW) to gather Reddit posts written by lonely authors in both the caregiver and non-caregiver populations. Our dataset consists of 15 subreddits. 8 "caregiver" subreddits that are related to caregivers or a diagnosis (r/AgingParents, r/cancer, r/CancerCaregivers, r/caregivers, r/caregiversofreddit, r/CaregiverSupport, r/dementia, r/DementiaHelp) and 7 "non-caregiver" subreddits (r/alone, r/ForeverAlone, r/loneliness, r/lonely, r/lonelywomen, r/mentalhealth, r/offmychest). 28,351 posts were scraped from the caregiver subreddits and 41,619 posts were scraped from the non-caregiver subreddits.

To explore the likelihood of cross-population contamination, 202 posts were randomly sampled from the non-caregiver subreddits and annotated for being written by a caregiver, with none found to be written by a caregiver, thus corroborating the "caregiver" subreddit and "non-caregiver" subreddit delineation.

\subsection{Data Preprocessing and Relevance Judging}

We applied a token count filter using the tiktoken package, Reddit posts below a token count of 150 and above 1000 were filtered out based on the hypothesis that excessively short posts were likely to lack enough textual evidence for meaningful analysis of loneliness while excessively long posts may have a diminished quality of writing. 

Following the token count filter, regular expression-based preprocessing for detecting irrelevant posts was applied to the caregiver subreddits' posts. Regular expression filtering was only utilized on the caregiver subreddits as their more limited range of topics compared to the non-caregiver subreddits allowed for the identification of specific phrases that indicated irrelevance or relevance. A student annotated sample (n=692) from the caregiver subreddits was annotated for whether or not the post's author was a caregiver resulted in subreddit-specific regular expressions. Regular expressions were chosen based on a qualitative investigation of this sample. For example, posts soliciting participation in research surveys were common in subreddits that lacked moderation and were able to be filtered out with keywords detecting the presence of weblinks or survey-related terms. Keywords that indicated the care recipient was in full-time care and thus not being cared for such as "in memory care" or "in assisted living" were also used as indicators of irrelevance. For disease oriented subreddits such as r/cancer and r/dementia, phrases that indicated the author was the patient such as "my cancer" or "I have cancer" were used as an indicator of irrelevance. For r/AgingParents, regex filtering for terms related to caregiver was used to detect relevant posts, as the subreddit was anticipated to have a lower proportion of caregiving related posts. 

Following preprocessing, relevant posts were determined via a relevance judging step with the aim of filtering out obviously irrelevant posts to reduce the dataset size that was ultimately run through the more complex and costly downstream steps. Thus, a low precision was considered acceptable for this step and we prioritized achieving a high recall. For posts from the caregiver subreddits, relevance was based on if a post was written by a caregiver, with GPT-4o achieving a precision of 0.971, recall of 0.937, and F1 score of 0.955 against a student annotated sample (n=692). For the non-caregiver subreddits, we applied GPT-5-nano to detect if the author was discussing their experience of loneliness, in first person, with textual evidence, achieving a precision of 0.583, recall of 1, and F1 score of 0.737 against a student annotator (n=202). 

\subsection{Loneliness Evaluation Framework}

To facilitate the creation of a loneliness evaluation framework, we built a preliminary dataset of Reddit posts that were judged to be written by a caregiver and scored highly on a prompt applying an unmodified UCLA Loneliness Scale. A sample of 29 high scoring posts was annotated by three co-author experts, professors of nursing with expertise in gerontology, loneliness, caregiver well-being, and technological/AI interventions, to inform the production of a modified version of the UCLA Loneliness scale. Based on this annotation process, we modified the scale's items such that it was more suitable for measuring the degree of loneliness from the perspective of a third party annotator reading open-ended Reddit data. Items were modified to be in the third person, such as "The author is unhappy doing so many things alone." The available answer choices for each item were simplified to "yes," indicating the presence of explicit textual evidence supporting the item, and "no," indicating the presence of evidence negating an item, with the addition of "not judgeable," representing a lack of relevant evidence, to account for the open-ended nature of Reddit. Detailed coding guidelines were introduced for each item that clarified the definition of each item, especially for application to the caregiver population. The coding guidelines included strict criteria for when to label "yes," "no," and "not judgeable" for each item along with examples of relevant topics to reduce ambiguity (See Table 4 in Appendix A.1). Additionally, the experts judged 5 of the original UCLA scale's items to be difficult to apply from the perspective of a third party and removed them from our modified scale.

We convert each label to a numeric score using the following scoring framework, with a high loneliness score indicating a higher degree of loneliness.

\[
\text{Item Score} =
\begin{cases}
\text{Yes} \mapsto 1, \\
\text{No} \mapsto -1, \\
\text{Not Judgeable} \mapsto 0.
\end{cases}
\]

\[
\text{Loneliness Score} = \sum_{i=1}^{15} \text{item}_i
\]

\subsection{Cause of Loneliness Categorization Framework}

Based on our literature survey and qualitative analysis of our collected Reddit posts, we created a cause of loneliness categorization framework that defines 7 distinct types of causes of loneliness. Our framework includes criteria for each type and guidelines that clarify each definition's verbiage with the aim of minimizing ambiguity between types (See Table 5 in Appendix A.2). This framework uses 2 axes: category of cause of loneliness and whether or not the cause is related to being a caregiver. Our framework was developed with review from a professor of nursing with expertise in loneliness, gerontology, and caregiving. 

\begin{enumerate}
\item \textbf{Social:} Dissatisfaction with quantity of relationships and/or frequency of social contact.

\item \textbf{Emotional:} Dissatisfaction with the closeness or intimacy of the author's relationship(s).

\item \textbf{Physical:} A loss of physical space, physical capacity, and/or physical and temporal boundaries.

\item \textbf{Mental Health:} A pre-existing mental health condition or state of poor mental health existent prior to the experience of loneliness.

\item \textbf{Relational:} A discrepancy in the author’s perception of themselves and their social network’s perception of the author’s identity, circumstances, or role.

\item \textbf{Network:} A sense of abandonment by the author's existing social network, often from a lack of physical labor or emotional support.

\item \textbf{Other:} The cause of the author’s loneliness is not fully described by any prior category.
\end{enumerate}

\subsubsection{Annotation Procedure}

Using the sample of 29 high scoring posts from the caregiver subreddits, we also took a proportional sample of 202 posts from the non-caregiver subreddits, applying the same score threshold to gather 31 high-scoring posts. Our aim was to gather 2 similarly-sized samples of posts written by lonely authors of both populations. Score thresholding was applied to prioritize prompting performance on texts exhibiting loneliness, versus texts with an absence of loneliness, and to validate our loneliness evaluation and cause categorization frameworks against texts that exhibited loneliness. The sample size of approximately 30 posts was chosen in order to reasonably limit annotation time and to prioritize high quality annotations over quantity. These two samples were annotated by a student annotator using the loneliness evaluation framework and the cause categorization framework to validate the data processing pipeline's prompts. For the demographic feature extraction step, 3 expert annotators independently labeled subsets of the caregiver sample (19, 19, and 20 posts, respectively), yielding 29 double-annotated posts across 9 demographic attributes, achieving an average Cohen's kappa of 0.84. These expert annotations were merged by a student annotator into a single ground truth dataset. 

\section{Results}
\subsection{Loneliness Evaluation Performance}

Following relevance judging, we applied the loneliness evaluation framework with GPT-5, using prompts that required the identification of explicit textual evidence, to measure the degree of loneliness exhibited by each post's author. 

\begin{table}[!h]
\centering
\footnotesize
\begin{tabular}{ccc}
\toprule
 & Caregiver (n=29) & Non-caregiver(n=31) \\
\cmidrule(lr){2-2} \cmidrule(lr){3-3}
Item 
& Accuracy (\%) 
& Accuracy (\%) \\
\midrule
1  & 86.21 & 93.55 \\
2  & 75.86 & 83.87 \\
3  & 82.76 & 83.87 \\
4  & 68.97 & 83.87 \\
5  & 68.97 & 83.87 \\
6  & 62.07 & 77.42 \\
7  & 72.41 & 74.19 \\
8  & 72.41 & 74.19 \\
9  & 75.86 & 67.74 \\
10 & 62.07 & 70.97 \\
11 & 65.52 & 67.74 \\
12 & 79.31 & 80.65 \\
13 & 75.86 & 77.42 \\
14 & 96.55 & 83.87 \\
15 & 96.55 & 93.55 \\
\midrule
Overall  & 76.09 & 79.78\\
\bottomrule
\end{tabular}
\caption{Loneliness evaluation accuracy by item for samples of the caregiver and non-caregiver subreddits.}
\end{table}

As presented in Table 1, GPT-5's average accuracy against the ground truth is relatively low at 76.09\% for the caregiver sample and 79.78\% for the non-caregiver sample. Accuracy was computed using exact label matching, with all incorrect predictions treated equally regardless of whether the label was “yes,” “no,” or “not judgeable.” Accuracy per label widely varied. In the caregiver sample, the accuracy ranged from 62.07\% to 96.55\%; the three best performing items were item 1, item 14 and item 15, in descending order. In the non-caregiver sample, the accuracy ranged from 67.74\% to 93.55\%; the three best performing items were item 1, item 15, and items 14, 2, 3, 4, 5 (See Table 4 in Appendix A.1 for and excerpt of the items). 

GPT-5 had the lowest accuracy on the "no" label in both samples (63.158\% accuracy in the caregiver sample, 46.667\% accuracy in the non-caregiver sample), with relatively high performance on the "yes" and "not judgeable labels (detailed results in Appendix A.5). 

We then applied this loneliness evaluation prompt to the dataset of Reddit posts, applying a total loneliness score threshold of 7, inclusive, to gather the posts written by authors experiencing a high degree of loneliness. Given the scoring range of -15 to +15, we chose 7 such that at least half of the 15 items of our framework should be able to be judged as "yes" using textual evidence from the post. 

\subsection{Cause of Loneliness Categorization Performance}

Following the loneliness score thresholding, GPT-5 was applied to the remaining posts to identify and categorize causes using the cause categorization framework. GPT-5's performance against human annotation is shown in Table 2. GPT-5 performed well, achieving a micro-aggregate F1 score of 0.825 in the caregiver sample and a micro-aggregate F1 score of 0.8 in the non-caregiver sample. Recall was higher in the caregiver sample than the non-caregiver sample, (0.87 versus 0.78 micro-aggregate recall), but precision was lower in the caregiver sample than the non-caregiver sample (0.784 versus 0.78 micro-aggregate precision). The macro-aggregate F1 score was much lower than the micro-aggregate F1 score for both samples, reflecting high performance on frequent classes but poorly on rare classes. 

Table 3 presents the results for identifying both the correct type of cause and whether it is related to caregiving, for the caregiver sample, as the non-caregiver sample did not contain any causes labeled as related to caregiving. We see lower performance, with a micro-aggregate F1 score of 0.721. 

\begin{table*}[h!]
\centering
\footnotesize
\begin{tabular}{l|cccccccccc}
\toprule
 & \multicolumn{5}{c}{Caregiver Sample (n=29) }
 & \multicolumn{5}{c}{Non-caregiver Sample (n=28)} \\
\cmidrule(lr){2-6} \cmidrule(lr){7-10}
Type & Precision & Recall & F1 & Accuracy & Count & Precision & Recall & F1 & Accuracy & Count \\
\midrule
Social       & 0.6 & 0.75 & 0.667 & 0.897 & 4& 0.92 & 1 & 0.958 & 0.929 & 23 \\
Emotional    & 0.667 & 0.8 & 0.727 & 0.897 &5& 0.75 & 0.857 & 0.8 & 0.786 & 14 \\
Physical      & 0.667 & 0.857 & 0.75 & 0.862 &7& 0 & 0 & 0 & 0.893 & 1\\
Mental Health & 0 & 0 & 0 & 0.931 &0& 1 & 0.333 & 0.5 & 0.714 &12\\
Relational  & 0.923 & 0.923 & 0.923 & 0.931 &13& 0.333 & 0.5 & 0.4 & 0.893 & 2\\
Network      & 0.938 & 0.882 & 0.909 & 0.897 &17& 1 & 0.857 & 0.923 & 0.964 &7\\
Other        & 0 & 0 & 0 & 1&  & 0 & 0 & 0 & 1 & 0\\
\midrule
Aggregate &  &  &  & 0.916 &46&  &  &  &  0.883 & 59\\
Macro-Agg. & 0.542 & 0.602 & 0.568 & && 0.571 & 0.507 & 0.512 \\
Micro-Agg. & 0.784 & 0.87 & 0.825 & && 0.821 & 0.78 & 0.8 &\\
\bottomrule
\end{tabular}
\caption{Metrics for cause of loneliness categorization on the caregiver and non-caregiver sample, with macro-aggregate and micro-aggregate metrics, rounded to the nearest thousandth. Due to an error, 28 of the 31 posts in the non-caregiver sample were utilized.}
\label{tab:noncaregiver-metrics}
\end{table*}

\begin{table}[h!]
\centering
\scriptsize
\begin{tabular}{l|ccccc}
\toprule
\makecell[l]{Type, Related\\ to Caregiving}  & Precision & Recall & F1 & Accuracy & Count  \\
\midrule
Social & 0.5 & 0.667 & 0.571 & 0.897 & 3\\
Emotional & 0 & 0 & 0 & 0.793 & 5 \\
Physical & 0.333 & 1 & 0.5 & 0.793 & 3 \\
Mental Health & 0 & 0 & 0 & 0.931  & 0\\
Relational & 0.923 & 0.857 & 0.889 & 0.897 & 14\\
Network & 0.933 & 0.824 & 0.875 & 0.862 & 17\\
Other          & 0 & 0 & 0 & 1 & 0\\
\midrule
\textbf{Aggregate}      &  &  &  &  0.882 \\
\textbf{Macro-Agg.} & 0.384 & 0.478 & 0.405 \\
\textbf{Micro-Agg.} & 0.705 & 0.738 & 0.721 \\
\bottomrule
\end{tabular}
\caption{Metrics for cause of loneliness categorization for identifying the correct type of cause and if a cause is related to the author being a caregiver, on the caregiver sample rounded to the nearest thousandth.}
\label{tab:caregiver-metrics}
\end{table}

\subsection{Demographic Extraction Performance}

We chose demographic attributes of interest that are related to the caregiving context to measure the overall diversity of the dataset at a macro-level. Following score thresholding, we used GPT-4o to extract 9 demographic attributes (caregiver gender,	caregiver age,	caregiving duration, patient gender, patient age, patient diagnosis, caregiver relationship with to patient, patient relationship with caregiver, relationship type) from the caregiver subreddits' posts. Against the ground truth, GPT-4o achieved an overall accuracy of 88.31\% (See Table 6 in Appendix A.2 for further details). 

\subsection{Final Dataset}

After applying our data processing pipeline of preprocessing, relevance judging, loneliness evaluation and loneliness score thresholding, 28,351 scraped posts from caregiver subreddits resulted in a final dataset of 387 posts written by lonely caregivers. From 41,619 scraped posts from the non-caregiver subreddits, a proportional sample of 4991 posts was fed into the data processing pipeline with the aim of achieving an approximately 1:2 ratio of the dataset sizes between caregivers and non-caregivers, for a final dataset of 908 posts written by lonely non-caregivers (See Table 7 and Table 8 in Appendix A.4 for details).

\subsection{Cause Distribution}
The cause of loneliness categorization framework was applied to both population's final datasets. Figure 1 shows the distribution of posts containing a cause of a given type, expressed as a proportion of all posts. We observe a distinct distribution of the types of causes of loneliness across the two populations. In posts written by caregivers, "network, related to caregiving" was the most widely found type of cause (56.1\% of posts) followed by "relational, related to caregiving" (37.13\%) "physical, related to caregiving" (29.54\%). In the non-caregiver dataset, the most commonly present types were "social, not related to caregiving"(74.19\%) and "emotional, not related to caregiving," (47.83\%) and "network, not related to caregiving" (23.58\%). Caregivers did have causes of loneliness unrelated to caregiving, with "social, not related to caregiving" (16.53\%) and "emotional, not related to caregiving" (7.32\%) being the most common. In both datasets, all types except other were present, when aggregating across related to or not related to caregiving.

\begin{figure*}[t]
    \centering
    \includegraphics[
        width=0.75\textwidth,
        trim={0cm 0cm 0cm 0cm}, clip
    ]{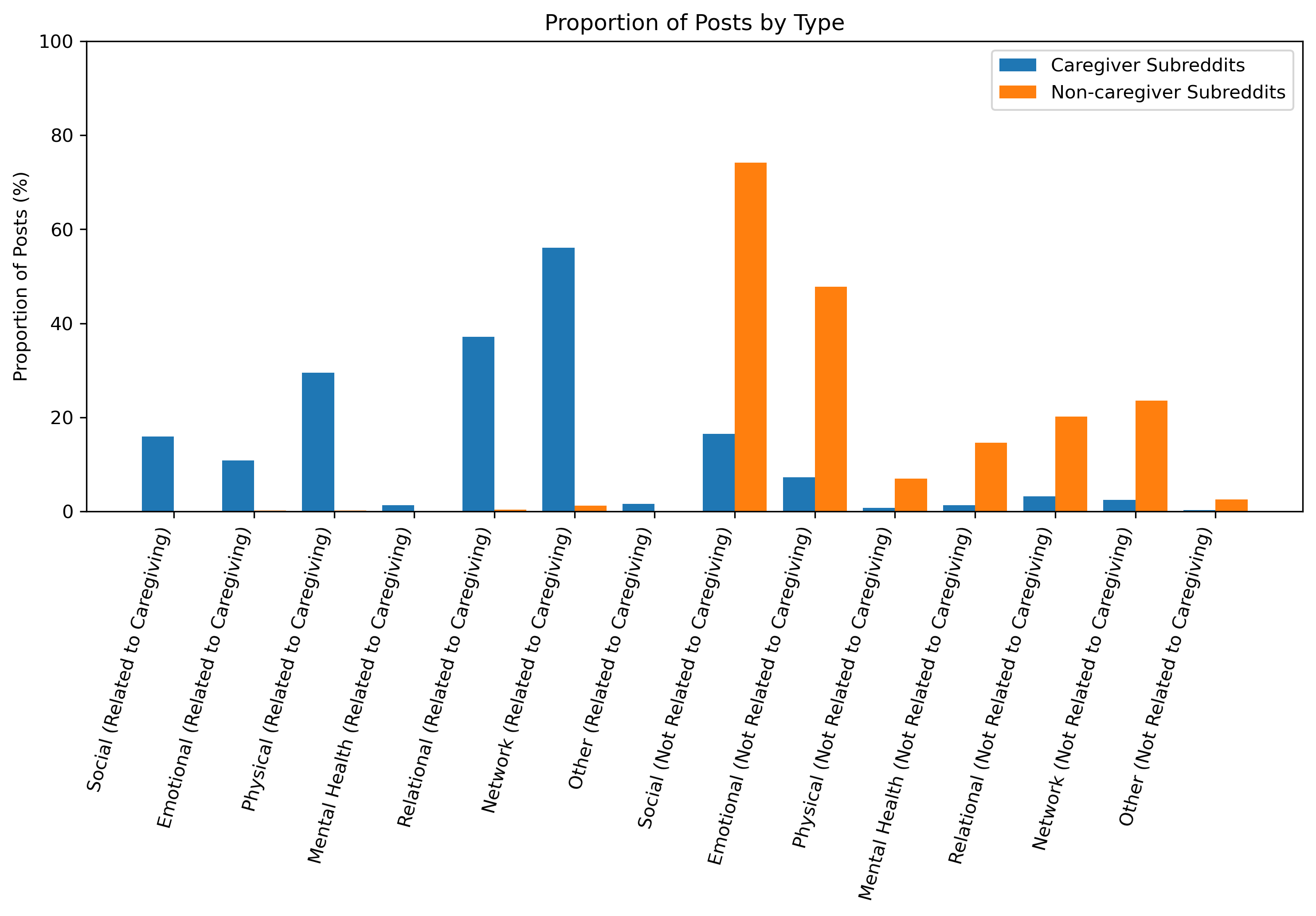}
    \caption{Proportion of all posts with a identified cause(s) of a given type in the caregiver and non-caregiver datasets.}
    \label{fig:prop_by_type}
\end{figure*}

\subsection{Demographic Extraction}

We applied our demographic extraction prompt to the caregiver loneliness dataset from an aggregate perspective. From the original 9 categories, 6 were used that were of interest to learn about the diversity of the caregiver dataset. Across 6 demographic attributes (caregiver age, caregiving duration, caregiver gender, caregiver relationship with patient, patient age, caregiver category based on patient diagnosis) the proportion of posts with known information for a given category averaged 50.267\% with caregiver gender being the least known (25.323\%) and caregiver relationship to patient being the most known (95.833\%). 

Out of the known posts, extracted demographic information was cleaned and categorized into bins. We can see that our dataset of high quality posts for the caregiver population is a diverse dataset across all 6 demographic categories, with all categorization bins for each demographic attribute having representation in the dataset (See Appendix A.6 for details). 

\section{Discussion and Analysis}

\subsection{Feasibility of Applying Psychological Frameworks to Reddit Data}
For the loneliness evaluation step, when selecting a loneliness threshold, there is a trade off between the dataset quality and the size of the dataset, especially for posts written by lonely caregivers. From the caregiver subreddits, only 387 posts out of 28,351 posts  (1.36\%) passed the score threshold of a total loneliness score greater than or equal to 7, indicating that the prevalence of high quality posts in the chosen caregiver subreddits is very low. As the posts were collected over 6 months, these subreddits are not fast-growing, indicating a ceiling on the amount of caregiver related posts available on Reddit. Thus, lowering the score threshold  may be necessary to increase the dataset size.

The aggregate accuracy (76.09\% for the caregiver sample and 79.78\% for the non-caregiver sample) suggests that GPT-5 is capable of accurately applying complex, psychologically grounded, annotation frameworks to measure loneliness based on social media text data.

For the model's performance, the GPT-5 had relatively high performance on the "yes" label (76.016\% accuracy on the caregiver sample versus 87.903\% accuracy on the non-caregiver sample) and "not judgeable" (79.47\% versus 72.777\%) labels, with lower performance on the "no" label (62.158\% versus 46.667\%) for both samples, suggesting the model is able to identify if textual evidence satisfies detailed coding criteria. The model is also able to accurately identify the absence of relevant textual evidence. However, when it comes to identifying textual evidence that negates an item of the loneliness evaluation scale, performance is notably lower (see Figure 2 and Figure 3 in Appendix A.5 for detailed confusion matrices). 

Our final dataset contains 387 posts written by lonely caregivers and 908 posts written by lonely non-caregivers that passed a score threshold of 7, meaning at least 7 out of 15 items on the loneliness scale were answered yes, with each item supported by exact, textual evidence. This demonstrates the feasibility of using Reddit as a scalable source of population-specific loneliness data, enabling the study of loneliness outside of interviewing or survey-based methods. 

In the non-caregiver sample, GPT-5's mean (7.17) and median (8) cumulative loneliness scores and the annotator's mean (7.59) and median (7) were similar. However, in the caregiver sample, the annotator's mean (7.17) was 1.48 greater than GPT-5's mean (5.69) and the annotator's median (8) was greater by 2 than GPT-5's median (6). As the prompt required providing explicit relevant evidence without any inference beyond the text, this suggests a discrepancy between what the annotator considered explicit, relevant evidence and the model. As the topic of the caregiver subreddits is not loneliness or mental health, there may be more ambiguous language that was explicit to the annotator, but not the model. Conversely, the annotator may have used more inference than the model. Thus, there may be a tradeoff between requiring explicit evidence and reducing inference for better model performance, and potentially missing evidence, resulting in fewer posts passing the loneliness threshold and a smaller final dataset. 

Of the extracted 9 demographic attributes, we analyzed 6. Information for a given demographic attribute was able to be identified in an average of 50.267\% of posts, suggesting Reddit users, despite the anonymous nature of the platform, are willing to share demographic information that is relevant to the subreddit topic, suggesting diversity can be examined via Reddit (see Table 9 in Appendix A.7 for per-attribute details). In the caregiver loneliness final dataset, all 6 demographic attributes have all of their categories represented in the dataset. We find that our dataset is diverse, suggesting our findings on population-level differences across caregiver and non-caregivers is informed by the experiences of a diverse range of caregiving contexts, supporting the robustness of our findings (See Appendix A.6 for detailed distributions of labels within each demographic attribute). Regarding specific attributes, for caregiver age, age tended to skew younger with 57.1\% of caregivers being age 30 or below, among known labels, potentially impacted by the social media context. A high proportion of caregivers were the child or grandchild of their care recipient(s) at 62.9\% of known labels. Care recipient's ages tended to be older, with 72.5\% of care recipients being 51 or older, among known labels.

Caregivers caring for patients with cancer and dementia were substantial proportions, at 27.5\% and 22.4\% of caregivers, among known labels, indicating that the contexts of dementia and cancer are not only significant groups in the caregiving population, but also significant groups when pertaining to caregiver loneliness. Additionally, diagnosis has a significant miscellaneous category at 40.1\%, demonstrating that while dementia and cancer are significant portions, our findings are based on a diverse representation of caregivers caring for many types of patients.

\subsection{Prompting for Loneliness Analysis}

For the cause categorization stage, we found several prompting strategies that improved performance. Firstly, the prompt included instructions to only consider current and explicitly stated causes of loneliness and to not make assumptions beyond the text in order to prevent misconstruing text that was generally related to loneliness as evidence for the presence of a specific type of cause of loneliness. This reduced the amount of inference and assumption the model would make, improving the precision at the expense of recall, as the model identified fewer but more accurate causes of the correct types.

To restrict the number of causes identified, the prompt also prohibited repeated usage of textual evidence across causes. This would encourage the causes to be distinct and reflect different types of causes opposed to several causes of the same type. To clarify causes of loneliness that may have some edge cases of overlap between types, the prompt included some prioritization rules. For example, sometimes the model would fail to recognize descriptions of lacking the time or energy to socialize as a cause of loneliness as authors were less likely to explicitly connect their loneliness and lack of leisure time. We added a rule that lacking time, energy, or physical capacity to engage socially and emotionally is eligible as a cause of loneliness and to prioritize labeling this cause as "physical". 

The model also tended to confuse the "relational" and "emotional" labels, as discussing an inability to relate was conflated with a lack of emotional closeness. Thus, we introduced a rule to not classify a cause as "emotional" unless the author explicitly discusses a lack of emotional closeness, to emphasize the key characteristics of the two types. Introducing examples of causes that must be categorized as "relational" such as misunderstanding, misperception, invalidation, or inaccurate role expectations also improved performance by clarifying how to decide between types, suggesting that reducing ambiguity between labels is important for accurate labeling. 

\subsection{Implications on Caregiver and Non-caregiver Experiences of Loneliness}

As shown in Figure 1, the distribution of types of causes differs substantially between the two populations. Less than 17\% of posts written by lonely caregivers contain a cause that is not related to caregiving, suggesting caregiving is a deeply involved role that impacts this population's experience of loneliness. 
Among caregivers, causes of loneliness are predominantly caregiving-related, indicating that caregiving roles and responsibilities are a primary driver of loneliness. In contrast to non-caregivers, with 74.3\% of posts containing a "social" cause (related to or not related to caregiving) and 48.05\% of posts containing an "emotional" cause, caregivers show relatively fewer causes related to relationship quantity or closeness ("social" 32.52\% and "emotional" 18.16\%), suggesting that their loneliness is less likely to be a result of desiring social network expansion or relationship deepening and more likely to be caused by unmet needs for support and recognition within their community ("relational" 40.38\% and "network" 40.38\%). 
As the cause types "relational" and "network" are based on themes of caregiver loneliness, this suggests the causes of caregiver loneliness are distinct from non-caregiver experiences of loneliness, and this is reflected in the final dataset and detected by the cause categorization prompt. These distinctions in the distribution of causes of loneliness across the 2 populations suggest that our dataset of Reddit posts is of sufficient quality to enable analysis of loneliness that reflects the specific characteristics of caregivers and non-caregivers, offering insight into the experiences of the two groups. For both populations, only 1.9\% and 2.56\% of posts had causes of the "other" label, suggesting that our cause categorization framework is comprehensive enough to be applied to most causes of loneliness, but complex enough to capture key differences between populations. 

\section{Conclusion and Future Works}

This work demonstrates the application of leveraging LLMs to build and analyze a diverse dataset of Reddit posts written by authors experiencing a high degree of loneliness in the caregiver and non-caregiver populations. Using a version of the UCLA loneliness scale modified for application to social media data and a psychologically grounded cause of loneliness categorization framework, we find that caregiver and non-caregiver loneliness are predominantly caused by distinct types of causes, with causes of caregiver loneliness being primarily related to caregiving. 

Building upon this work, future works to improve the data processing pipeline and explore more differences across the two populations are viable.

1. \textbf{Validating prompts against more robust annotation.} Double annotating the high-quality posts sample and building a gold annotated dataset would make the data processing pipeline's prompts more robust and identify more potential areas of ambiguity in the existing annotation guidelines. Increasing the number of posts annotated to 50 or 100 posts per population would be a stronger source of validation for the prompts utilized in this work. As this work's annotator was a student, using expert annotators would make the annotations a more robust source of truth.

2. \textbf{Experimenting with varied LLM models and open source models.} This work utilized GPT-5, GPT-5-nano, and GPT-4o. Introducing more models, such as Gemini, Deepseek or Claude could improve model performance, particularly on the loneliness evaluation task. Testing a voting scheme between models could additionally improve performance.

3. \textbf{Expanding data sources.} Expanding sources of data, particularly for the caregiver population, to additional subreddits or beyond the Reddit context could allow for an even more diverse and a larger final dataset. 

4. \textbf{Analyzing Additional Characteristics} Applying our datasets to gain further insight, particularly into the treatment of loneliness, is a strong future direction. Future studies should leverage the final datasets to analyze additional characteristics of loneliness, such as symptom distribution, to work toward applying the dataset to inform intervention.

\section*{Limitations}

\textbf{Unbalanced Samples.} Despite strong performance in the micro-aggregate F1 score in the cause categorization experiment, the low macro-aggregate F1 score for both populations suggests the model performed well on frequent classes, but poorly on classes that were underrepresented in the annotated sample. One limitation of the cause categorization experiment is the unbalanced representation of each type of cause in the samples used to prompt engineer, which may have favored prompts that performed well on types with greater representation. 

\textbf{Annotation Methodology.} Another limitation is our usage of a single undergraduate student as the annotator opposed to multiple experts for the loneliness evaluation and cause categorization stages. Our prompt may be overfit to the biases or annotation style of the single annotator. Using expert annotators and/or multiple annotators would make our prompt's performance more robust. 

\textbf{Limited Annotated Data.} The small size of our annotated data may limit the validation of our prompts, as testing prompts against a larger set of annotated data, such as 50 to 100 posts per population, would strengthen confidence in our prompts' performance when applied to large datasets of open-ended textual data. 

\section*{Ethical Considerations}

\textbf{IRB Status and Dataset Availability.} Currently, this project has not undergone Institutional Review Board (IRB) consultation. Thus, our dataset will not be made publically available in order to prevent the identification of the Reddit posts utilized in this study. 

\textbf{Anonymization of Data.} In order to anonymize our data, we removed metadata such as usernames, time of post creation, time of post update, number of upvotes, upvote ratio, and over 18 rating. All experiments were conducted using solely the post's textual content. We collected only textual content necessary for the research questions and removed all metadata not required for analysis. Prior to experiments, we stripped references to usernames from the posts' content.  We solely attempt to learn about loneliness on a population level and do not attempt to identify individuals. No verbatim quotations of Reddit posts are included in this paper. We do not discuss or paraphrase specific posts in this paper and strictly only discuss general themes on a dataset level, without referring to any specific instance of data. When discussing demographics, we additionally only discuss the overall dataset and share aggregated results. 

\textbf{Utilizing Social Media Data for Mental Health Analysis.} This work collects and analyses anonymous, publically visible, Reddit posts to learn about loneliness in caregiver and non-caregivers. We utilized all data in compliance with Reddit's Terms of Service, API Policies, and ethical research standards. We did not contact any user or moderator of the subreddits we utilized at any point of this work.

\textbf{Misapplication of Findings.} While this work presents a foundational step toward utilizing LLMs to measure and analyze loneliness, we do not recommend direct application of this work on humans. Our work aims to assess the capability of LLMs and learn on a population level, not to be used on individuals for the purpose of assessing individual loneliness as there is significant risk of misjudging the presence, extent of, and manifestation of loneliness. Our work is not intended to and should not be applied to replace professional mental health help.

\section*{Acknowledgments}

We are grateful to the reviewers for their valuable insights and constructive feedback that greatly improved this work. In addition, we thank the Emory University Computer Science Department and our colleagues at the Emory NLP for their technical support and computational resources that made this work possible. 

\bibliography{custom}

@inproceedings{national-alliance-2025,
    title = "{Caregiving in the US 2025}",
    author = "AARP and National Alliance for Caregiving",
    booktitle = "Alzheimer's \& Dementia",
    month = jul,
    year = "2025",
    address = "Washington, DC",
    publisher = "",
    url = "",
    doi = "10.26419/ppi.00373.001",
    pages = "1--129",
}

@article{vasileiou-2017,
    title = "{Experiences of Loneliness Associated with Being an Informal Caregiver: A Qualitative Investigation}",
    author = "Vasileiou, Konstantia and
        Barnett, Julie and
        Barreto, Manuela and
        Vines, John and
        Atkinson, Mark and
        Lawson, Shaun and
        Wilson, Michael",
    journal = "Frontiers in Psychology",
    volume = "8",
    articleno = {585},
    month = apr,
    year = "2017",
    issn = "1664-1078",
    address = "",
    publisher = "Frontiers in Psychology",
    url = "https://www.frontiersin.org/journals/psychology/articles/10.3389/fpsyg.2017.00585",
    doi = "10.3389/fpsyg.2017.00585",
    pages = "",
}

@article{victor-2021,
    author = {Christina R. Victor and Isla Rippon and Catherine Quinn and Sharon M. Nelis and Anthony Martyr and Nicola Hart and Ruth Lamont and Linda Clare},
    title = {The prevalence and predictors of loneliness in caregivers of people with dementia: findings from the IDEAL programme},
    journal = {Aging \& Mental Health},
    volume = {25},
    number = {7},
    pages = {1232--1238},
    month = apr,
    year = {2020},
    publisher = {Routledge},
    doi = {10.1080/13607863.2020.1753014},
    note ={PMID: 32306759},
    URL = {https://doi.org/10.1080/13607863.2020.1753014},
    eprint = {https://doi.org/10.1080/13607863.2020.1753014}
}

@article{mcrae-2009,
	title = {Predictors of loneliness in caregivers of persons with Parkinson's disease},
	volume = {15},
	issn = {1353-8020},
	url = {https://www.sciencedirect.com/science/article/pii/S135380200900042X},
	doi = {https://doi.org/10.1016/j.parkreldis.2009.01.007},
	pages = {554--557},
	number = {8},
	journaltitle = {Parkinsonism \& Related Disorders},
	author = {{McRae}, Cynthia and Fazio, Emily and Hartsock, Gina and Kelley, Livia and Urbanski, Shawna and Russell, Dan},
	date = {2009},
}

@inproceedings{yang-2023,
    title = "Towards Interpretable Mental Health Analysis with Large Language Models",
    author = "Yang, Kailai  and
      Ji, Shaoxiong  and
      Zhang, Tianlin  and
      Xie, Qianqian  and
      Kuang, Ziyan  and
      Ananiadou, Sophia",
    editor = "Bouamor, Houda  and
      Pino, Juan  and
      Bali, Kalika",
    booktitle = "Proceedings of the 2023 Conference on Empirical Methods in Natural Language Processing",
    month = dec,
    year = "2023",
    address = "Singapore",
    publisher = "Association for Computational Linguistics",
    url = "https://aclanthology.org/2023.emnlp-main.370/",
    doi = "10.18653/v1/2023.emnlp-main.370",
    pages = "6056--6077"
}

@article{jiang-2022,
	title = {Many Ways to Be Lonely: Fine-Grained Characterization of Loneliness and Its Potential Changes in {COVID}-19},
	volume = {16},
	url = {https://ojs.aaai.org/index.php/ICWSM/article/view/19302},
	doi = {10.1609/icwsm.v16i1.19302},
	pages = {405--416},
	number = {1},
	journaltitle = {Proceedings of the International {AAAI} Conference on Web and Social Media},
	shortjournal = {{ICWSM}},
	author = {Jiang, Yueyi and Jiang, Yunfan and Leqi, Liu and Winkielman, Piotr},
	urldate = {2025-11-04},
	date = {2022-05-31},
	note = {Section: Full Papers},
}

@inproceedings{fujikawa-2024,
    title = "Loneliness Episodes: A {J}apanese Dataset for Loneliness Detection and Analysis",
    author = "Fujikawa, Naoya  and
      Toan, Nguyen  and
      Ito, Kazuhiro  and
      Wakamiya, Shoko  and
      Aramaki, Eiji",
    editor = "De Clercq, Orph{\'e}e  and
      Barriere, Valentin  and
      Barnes, Jeremy  and
      Klinger, Roman  and
      Sedoc, Jo{\~a}o  and
      Tafreshi, Shabnam",
    booktitle = "Proceedings of the 14th Workshop on Computational Approaches to Subjectivity, Sentiment, {\&} Social Media Analysis",
    month = aug,
    year = "2024",
    address = "Bangkok, Thailand",
    publisher = "Association for Computational Linguistics",
    url = "https://aclanthology.org/2024.wassa-1.23/",
    doi = "10.18653/v1/2024.wassa-1.23",
    pages = "280--293"
}

@article{gierveld-1998,
  author={Gierveld, Jenny de Jong}, 
  title={A review of loneliness: concept and definitions, determinants and consequences},
  journal={Reviews in Clinical Gerontology}, 
  volume={8}, 
  number={1},   
  pages={73–80},
  DOI={10.1017/S0959259898008090},
  url={https://doi.org/10.1186/s13643-018-0935-y},
  year={1998}
}

@article{bonin-2021,
    author = {Bonin-Guillaume, Sylvie and Arlotto, Sylvie and Blin, Alice and Gentile, Stéphanie},
    year = {2022},
    month = {06},
    pages = {7050},
    title = {Family Caregiver’s Loneliness and Related Health Factors: What Can Be Changed?},
    volume = {19},
    journal = {International Journal of Environmental Research and Public Health},
    doi = {10.3390/ijerph19127050}
}

@article{gray-2019,
    author = {Gray, Tamryn and Azizoddin, Desiree and Nersesian, Paula},
    year = {2019},
    month = {10},
    pages = {1-9},
    title = {Loneliness among cancer caregivers: A narrative review},
    volume = {18},
    journal = {Palliative and Supportive Care},
    doi = {10.1017/S1478951519000804}
}

@article{russell-1978,
    author = {Russell, Daniel and Peplau, Letitia and Ferguson, Mary},
    year = {1978},
    month = {07},
    pages = {290-4},
    title = {Developing a Measure of Loneliness},
    volume = {42},
    journal = {Journal of personality assessment},
    doi = {10.1207/s15327752jpa4203_11}
}

@article{gierveld-2006,
    author = {Jenny De Jong Gierveld and Theo Van Tilburg},
    title ={A 6-Item Scale for Overall, Emotional, and Social Loneliness: Confirmatory Tests on Survey Data},
    journal = {Research on Aging},
    volume = {28},
    number = {5},
    pages = {582-598},
    year = {2006},
    doi = {10.1177/0164027506289723},
    URL = {https://doi.org/10.1177/0164027506289723},
    eprint = {https://doi.org/10.1177/0164027506289723}
}

@article{gierveld-1985,
    author = {Jenny de Jong-Gierveld and Frans Kamphuls},
    title ={The Development of a Rasch-Type Loneliness Scale},
    journal = {Applied Psychological Measurement},
    volume = {9},
    number = {3},
    pages = {289-299},
    year = {1985},
    doi = {10.1177/014662168500900307},
    URL = {https://doi.org/10.1177/014662168500900307},
    eprint = {https://doi.org/10.1177/014662168500900307}
}

@article{russell-1984,
    author = {Russell, Daniel and Cutrona, Carolyn and Rose, Jayne and Yurko, Karen},
    year = {1984},
    month = {06},
    pages = {1313-1321},
    title = {Social and emotional loneliness: An examination of Weiss's typology of loneliness},
    volume = {46},
    journal = {Journal of Personality and Social Psychology},
    doi = {10.1037/0022-3514.46.6.1313}
}

@article{motta-2021,
    author = {Motta, Valeria},
    year = {2021},
    month = {01},
    pages = {71-81},
    title = {Key Concept: Loneliness},
    volume = {28},
    journal = {Philosophy, Psychiatry, \& Psychology},
    doi = {10.1353/ppp.2021.0012}
}

@inproceedings{garg-2023,
author = {Garg, Muskan and Saxena, Chandni and Samanta, Debabrata and Dorr, Bonnie J.},
title = {LonXplain: Lonesomeness as\&nbsp;a\&nbsp;Consequence of\&nbsp;Mental Disturbance in\&nbsp;Reddit Posts},
year = {2023},
isbn = {978-3-031-35319-2},
publisher = {Springer-Verlag},
address = {Berlin, Heidelberg},
url = {https://doi.org/10.1007/978-3-031-35320-8_27},
doi = {10.1007/978-3-031-35320-8_27},
booktitle = {Natural Language Processing and Information Systems: 28th International Conference on Applications of Natural Language to Information Systems, NLDB 2023, Derby, UK, June 21–23, 2023, Proceedings},
pages = {379–390},
numpages = {12},
location = {Derby, United Kingdom}
}

\appendix
\section{Appendix}
\label{sec:appendix}

\subsection{Loneliness Evaluation Framework, Abbreviated}

Table 4 presents an abbreviated version of the loneliness evaluation framework. All 15 items have coding guidelines that were applied in the loneliness evaluation prompt. 

\begin{table}[htbp]
\centering
\footnotesize
\begin{tabular}{|p{0.5cm}|p{2cm}|p{3.5cm}|}
\hline
\textbf{Item} & \textbf{Item} & \textbf{Coding Guidelines} \\
\hline
1 & The author is unhappy doing so many things alone. & 
To rate this as "yes," they clearly express feeling alone and unhappy, often mentioning multiple activities or situations they are facing alone in the given context. \\
\hline
14 & It is difficult for the author to make friends. & This item pertains to evidence that the caregiver is attempting to reach out and make new connections. Existing friendships or family relationship changes are irrelevant. Connecting with other caregivers is relevant. \\
\hline
15 & People are around the author but not with them. & Two issues: does the author have people in their lives, and are they supportive? Being around: physically/socially present. Being “with”: spending quality time, alignment with caregiving approach, and/or feeling emotionally close. \\
\hline
\end{tabular}
\caption{Loneliness evaluation items and their corresponding coding guidelines.}
\label{tab:annotation-guidelines}
\end{table}

\subsection{Cause of Loneliness Categorization Framework, Abbreviated}

Table 5 presents an abbreviated version of the cause categorization framework, demonstrating the criteria for a cause to qualify as a given type and the additional guidelines given to clarify each type of cause. 

\begin{table}[htbp]
\centering
\footnotesize
\begin{tabular}{|p{1.25cm}|p{2cm}|p{3cm}|}
\hline
\textbf{Type} & \textbf{Criteria} & \textbf{Additional Guidelines} \\
\hline
Social & The author’s cause of
loneliness MUST be
dissatisfaction with their
quantity of relationships
and/or frequency of social
contact. & 
Often mentioning a lack of people
around them or lacking a social network.
“Around them” can include physical distance, such as a dissatisfaction with the number of friends nearby. \\
\hline
Emotional & The author’s cause of
loneliness MUST be
dissatisfaction with the
closeness or intimacy of
their relationship(s). & 
Pertaining to a lack of close and intimate relationships.
Often mentioning loneliness as a result
of an actual and/or perceived loss of
a close relationship (death can also be
considered the loss of a relationship).
Often mentioning loneliness as a result
of a desire for a close or intimate relationship and a perceived inability to
fulfill that desire.\\
\hline
\end{tabular}
\caption{Annotation items and corresponding coding guidelines.}
\label{tab:annotation-guidelines}
\end{table}

\subsection{Demographic Extraction Performance}

Table 6 presents GPT-4o's accuracy for extracting 9 demographic attributes, validated against expert annotation of a sample (n=29) from the caregiver subreddits.

\begin{table}[htbp]
\centering
\footnotesize
\begin{tabular}{l|c}
\toprule
Label & Accuracy (\%) \\
\midrule
Caregiver Gender        & 89.66 \\
Caregiver Age           & 91.38 \\
Caregiving Duration     & 84.48 \\
Patient Gender          & 94.83 \\
Patient Age             & 96.55 \\
Diagnosis                & 84.48 \\
\makecell[l]{Caregiver to Patient\\ Relationship}  & 91.38 \\
\makecell[l]{Patient to Caregiver\\ Relationship}   & 65.52 \\
Relationship Type       & 96.55 \\
\midrule
Overall & 88.31 \\
\bottomrule
\end{tabular}
\caption{The accuracy of GPT-4o on the caregiver subreddit sample(n=29) against the expert demographic annotation for each demographic label.}
\label{tab:annotation-accuracy}
\end{table}

\subsection{Data Processing Pipeline to Final Dataset, by Subreddit}

Table 7 and Table 8 present the number of posts at each stage of the data processing pipeline, by subreddit, until the final dataset of posts written by authors experiencing a high degree of loneliness.

\begin{table}[h!]
    \centering
    \scriptsize
    \begin{tabular}{c|ccccc}
         \toprule
         Subreddit&  Total & Relevance & Loneliness $\geq$ 7\\
         \midrule
         AgingParents& 5391&     355& 9\\
         cancer&  5346&  401& 1\\
         CancerCaregivers& 1578& 970& 41\\
         caregivers& 1307&   548& 27\\
         caregiversofreddit& 271& 41& 2\\
         CaregiverSupport&  5957&   3407& 236\\
         dementia& 8014&  3154& 68\\
         DementiaHelp& 487& 110& 3\\
         \midrule
         Total & 28351 & 8631 & 387\\
         \bottomrule
    \end{tabular}
    \caption{Post counts after each stage of the pipeline for the caregiver subreddits, resulting in the final dataset.}
    \label{}
\end{table}

\begin{table}[htbp]
    \centering
    \scriptsize
    \begin{tabular}{c|cccc}
         \toprule
         Subreddit&  Total &  Sample & Relevance & Loneliness $\geq$ 7\\
         \midrule
         alone& 2644 & 238& 217& 112\\
         ForeverAlone& 1473 & 152&  122& 32\\
         loneliness& 3132 & 320&  279& 101\\
         lonely& 9723 & 839&  783& 298\\
         lonelywomen& 348 & 28&  23& 10\\
         mentalhealth& 17377 & 2333& 857& 274\\
         offmychest& 6994 & 1081&  399& 81\\
         \midrule
         Total & 41619 & 4991 &  2680& 908\\
         \bottomrule
    \end{tabular}
    \caption{Post counts after each stage of pipeline for the non-caregiver subreddits, resulting in the final dataset.}
    \label{}
\end{table}

\subsection{Loneliness Evaluation Confusion Matrices}

Figure 2 and Figure 3 present the confusion matrices for the loneliness evaluation experiment, for the caregiver and non-caregiver samples respectively, with the proportion of each type of label that were labeled as "yes," "no," or "not judgeable" by GPT-5.

\begin{figure}[htbp]
    \centering
    \includegraphics[width=\columnwidth]{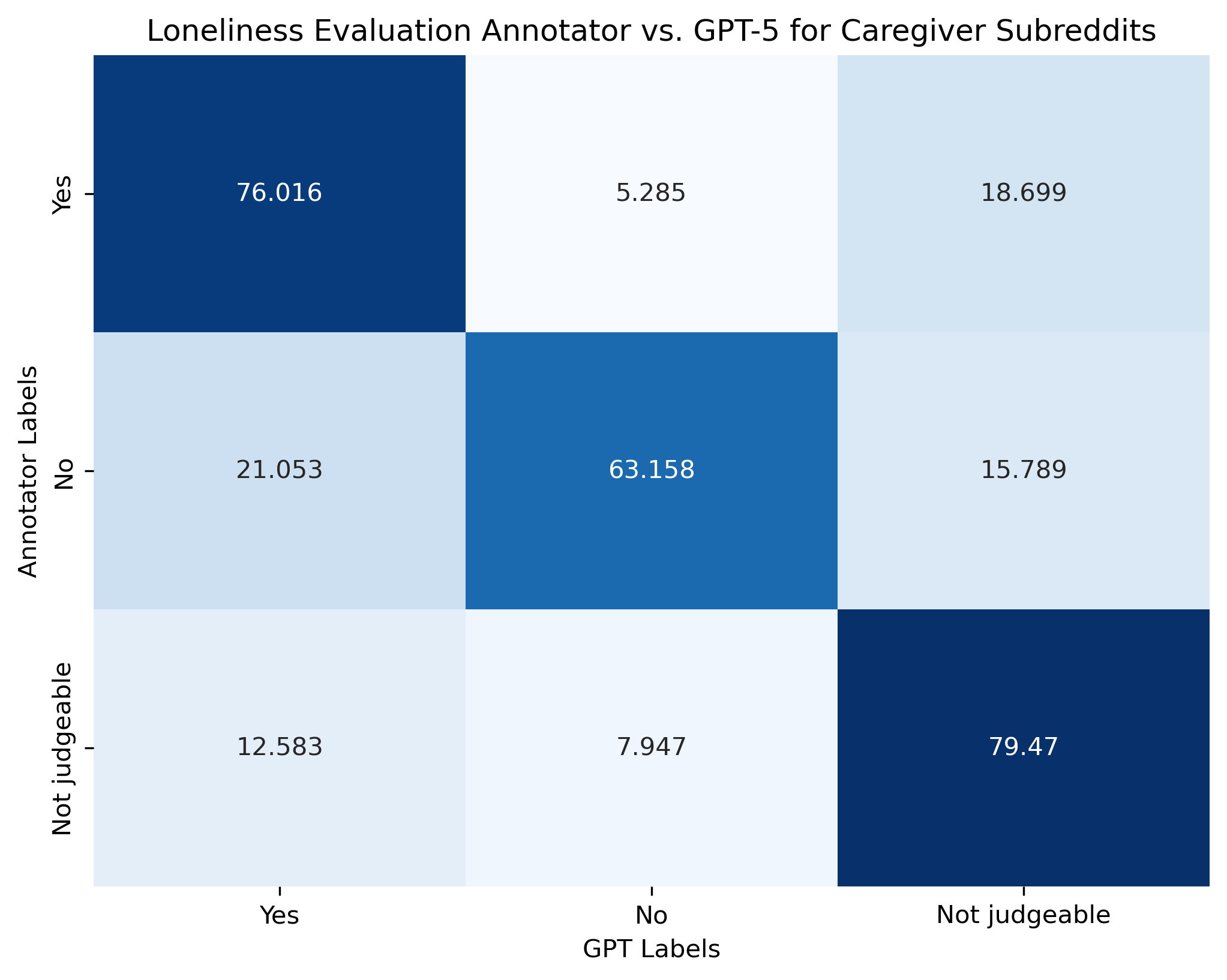}
    \caption{Confusion matrix showing the aggregate accu- racy of GPT-5 for answering the loneliness evaluation framework’s items for the caregiver subreddit sample. Depicts out of all annotator’s labels of a certain label, what proportion was labeled as each label by GPT-5.}
    \label{fig:placeholder}
\end{figure}

\begin{figure}[h!]
    \centering
    \includegraphics[width=\columnwidth]{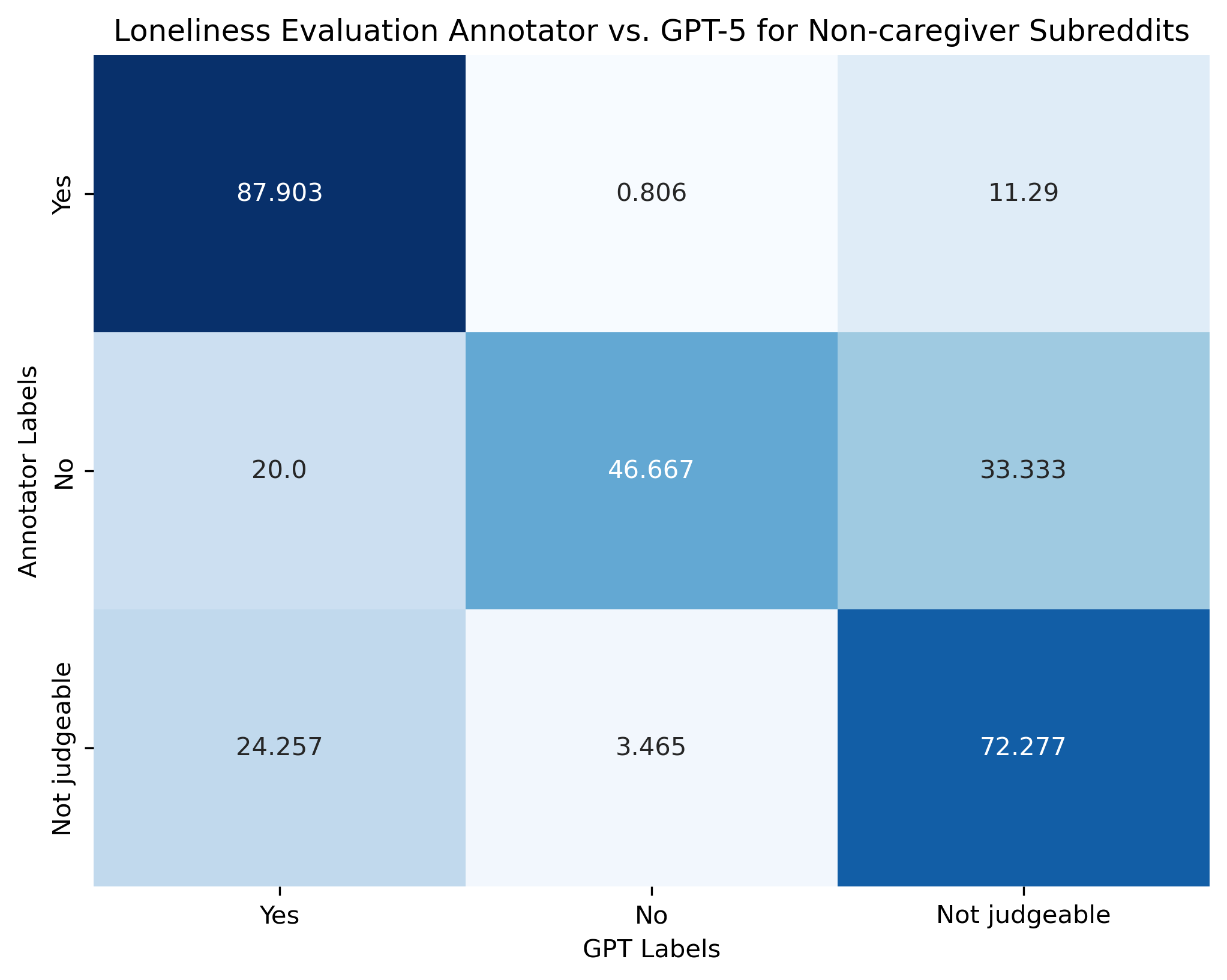}
    \caption{Confusion matrix showing the aggregate accu- racy of GPT-5 for answering the loneliness evaluation framework’s items for the non-caregiver subreddit sam- ple. Depicts out of all annotator’s labels of a certain label, what proportion was labeled as each label by GPT- 5.}
    \label{fig:placeholder}
\end{figure}

\subsection{Distribution of Demographic Attributes}
To examine if our dataset of lonely caregivers is diverse, we extracted demographic information using GPT-4o. Demographic information was cleaned and categorized into bins for each attribute. Demographic attributes were selected for relevance to a caregiving context, considering the likelihood of disclosure given the anonymous and public nature of Reddit.

\subsubsection{Caregiver Age}

Caregiver age is divided into bins of 10 years each, from age 11 to 80. 

\begin{figure}[h!]
    \centering
\includegraphics[width=\columnwidth, trim={0cm 0cm 0cm 0cm}, clip]{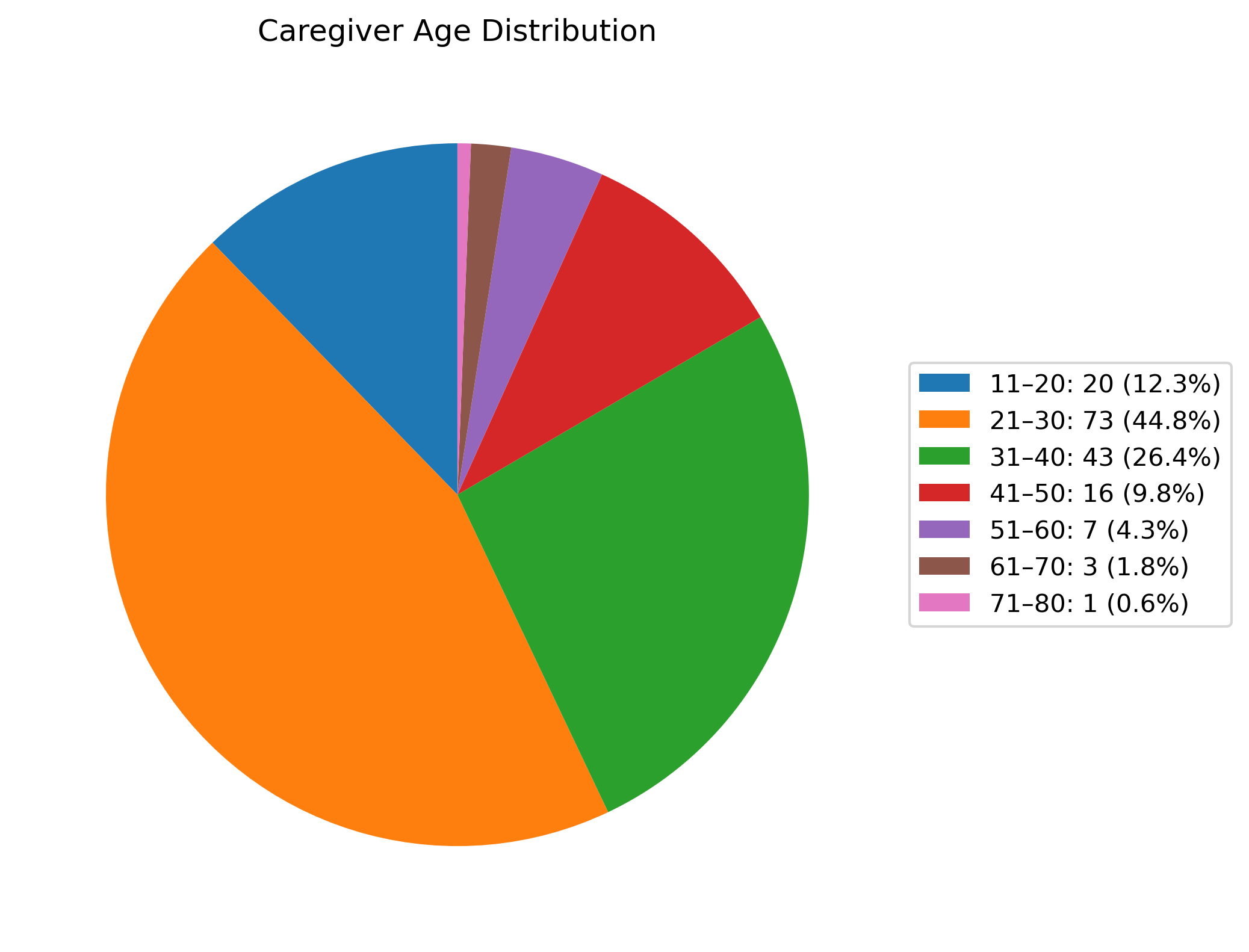}
\label{fig:placeholder}
    \caption{Distribution of posts, by caregiver age, among known labels.}
\end{figure}

\subsubsection{Caregiving Duration}

Caregiver duration is divided into less than a year, 1 to 5 years, 5 to 10 years, 10 to 20 years, and 20 to 30 years. These bins were chosen based on qualitative examination of the dataset.

\begin{figure}[h!]
    \centering
\includegraphics[width=\columnwidth, trim={0cm 0cm 0cm 0cm}, clip]{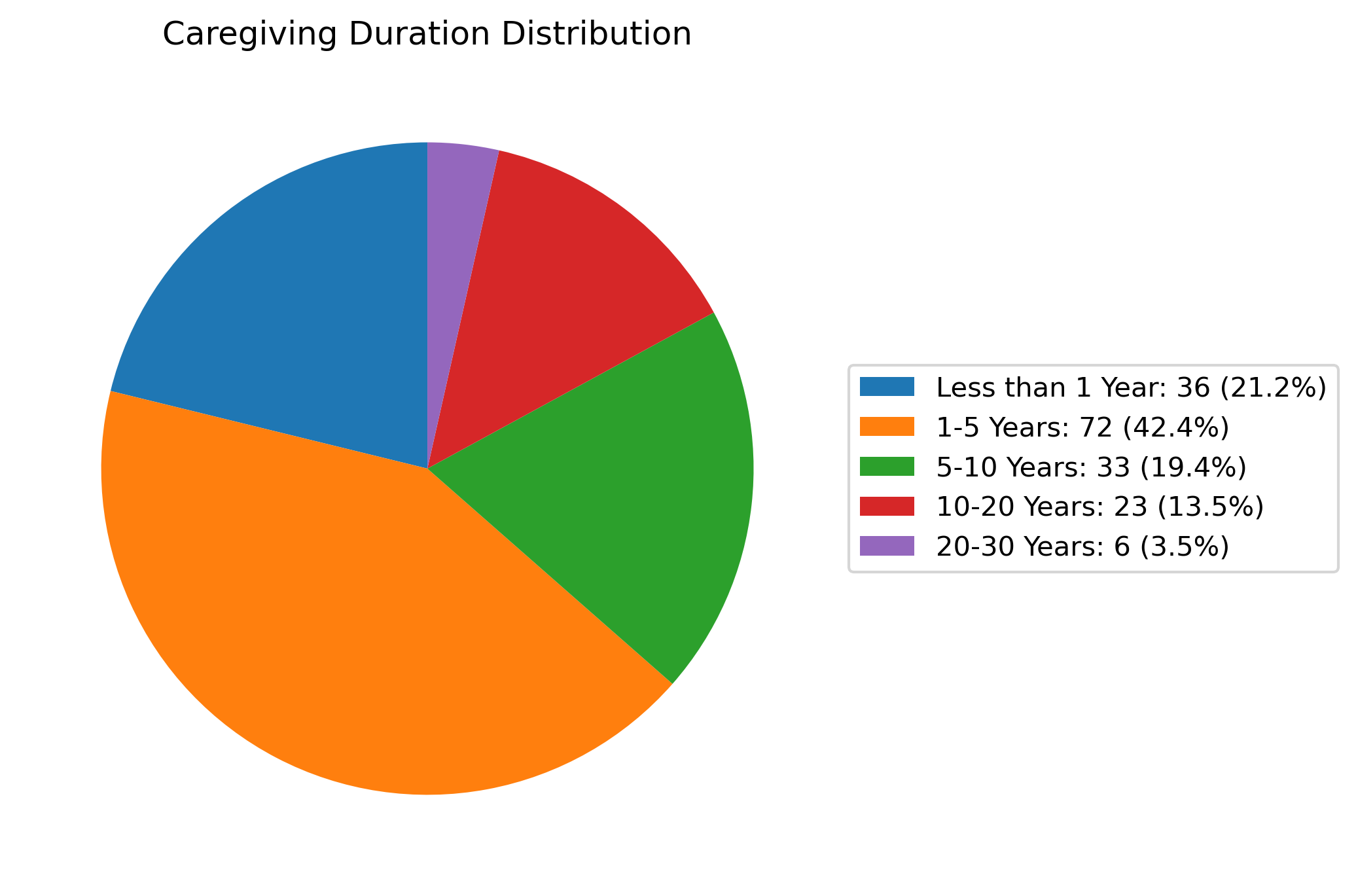}
\label{fig:placeholder}
    \caption{Distribution of posts, by caregiving duration, among known labels.}
\end{figure}

\subsubsection{Caregiver Gender}

Caregiver gender is heavily female (85.7\%) with male caregivers (14.3\%) being present in the dataset.

\begin{figure}[h!]
    \centering
\includegraphics[width=\columnwidth, trim={0cm 0cm 0cm 0cm}, clip]{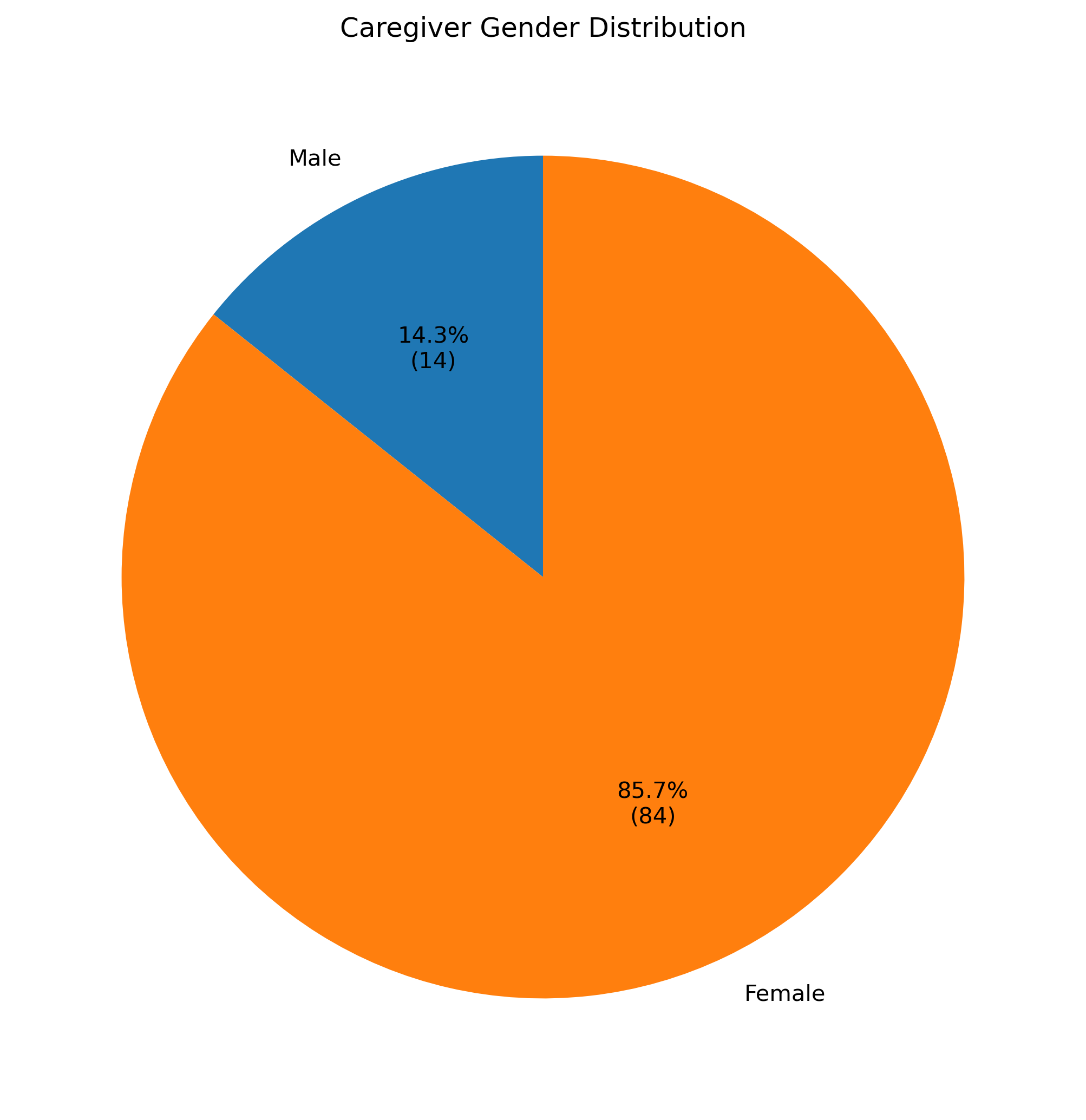}
\label{fig:placeholder}
    \caption{Distribution of posts, by caregiving gender, among known labels.}
\end{figure}

\subsubsection{Caregiver Relationship with Patient, per Patient}

As caregivers may care for more than one care recipient, we extracted the relationship of the caregiver with the patient, from the perspective of the caregiver, for each patient (n=456). 
\begin{figure}[h!]
    \centering
\includegraphics[width=\columnwidth, trim={0cm 0cm 0cm 0cm}, clip]{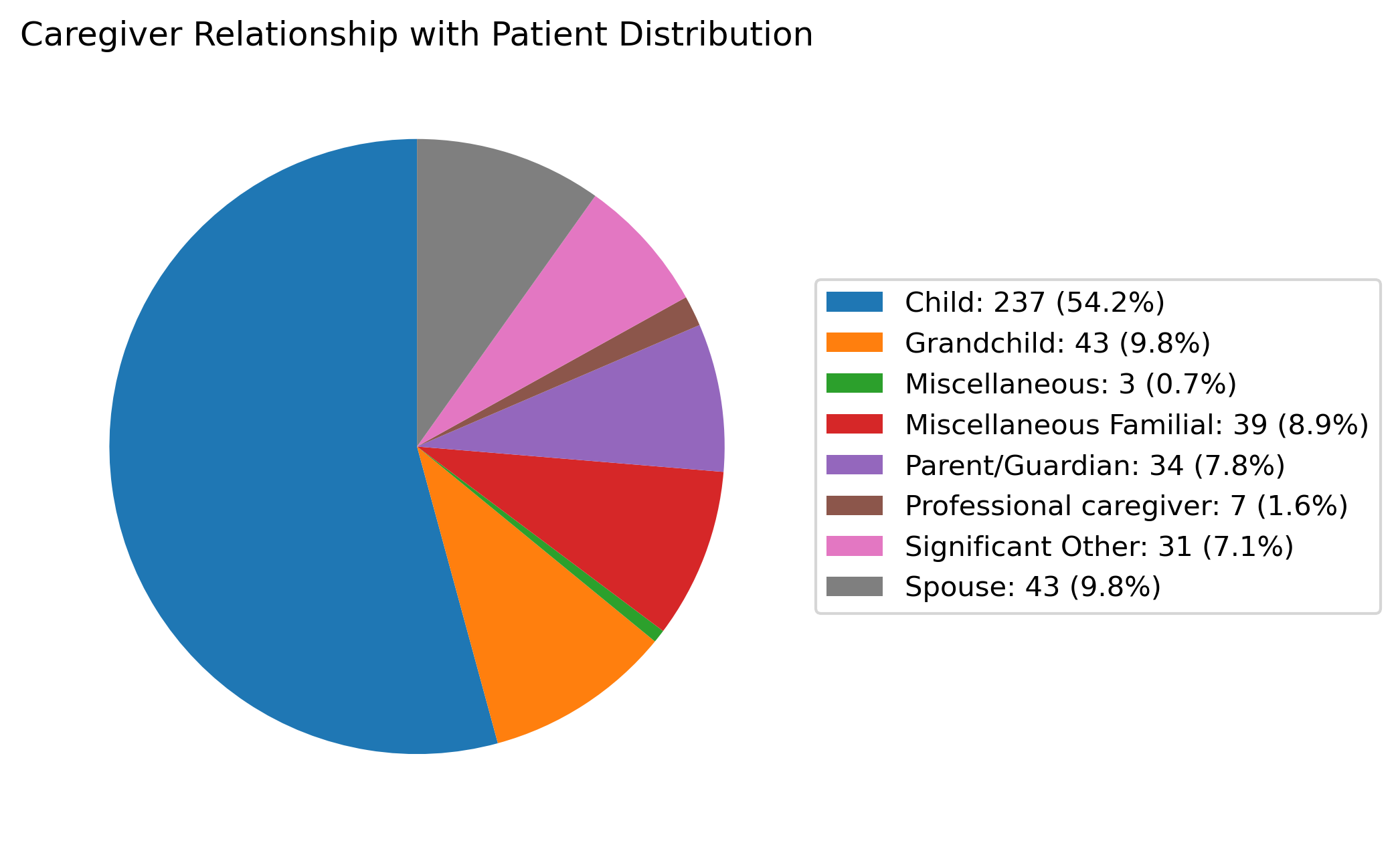}
\label{fig:placeholder}
    \caption{Distribution of caregiver relationship to patient among known labels, by patient.}
\end{figure}

\subsubsection{Patient Age}
Patient age is also divided in 10 year increments, starting at 0 years of age to over 91 years of age. Older ages are more prevalent with 20.1\% of patients being ages 71 through 80 and an additional 20.1\% being ages 81 through 90. However, all age bins are represented in the final caregiver loneliness dataset.
\begin{figure}[h]
    \centering
\includegraphics[width=\columnwidth,, trim={0cm 0cm 0cm 0cm}, clip]{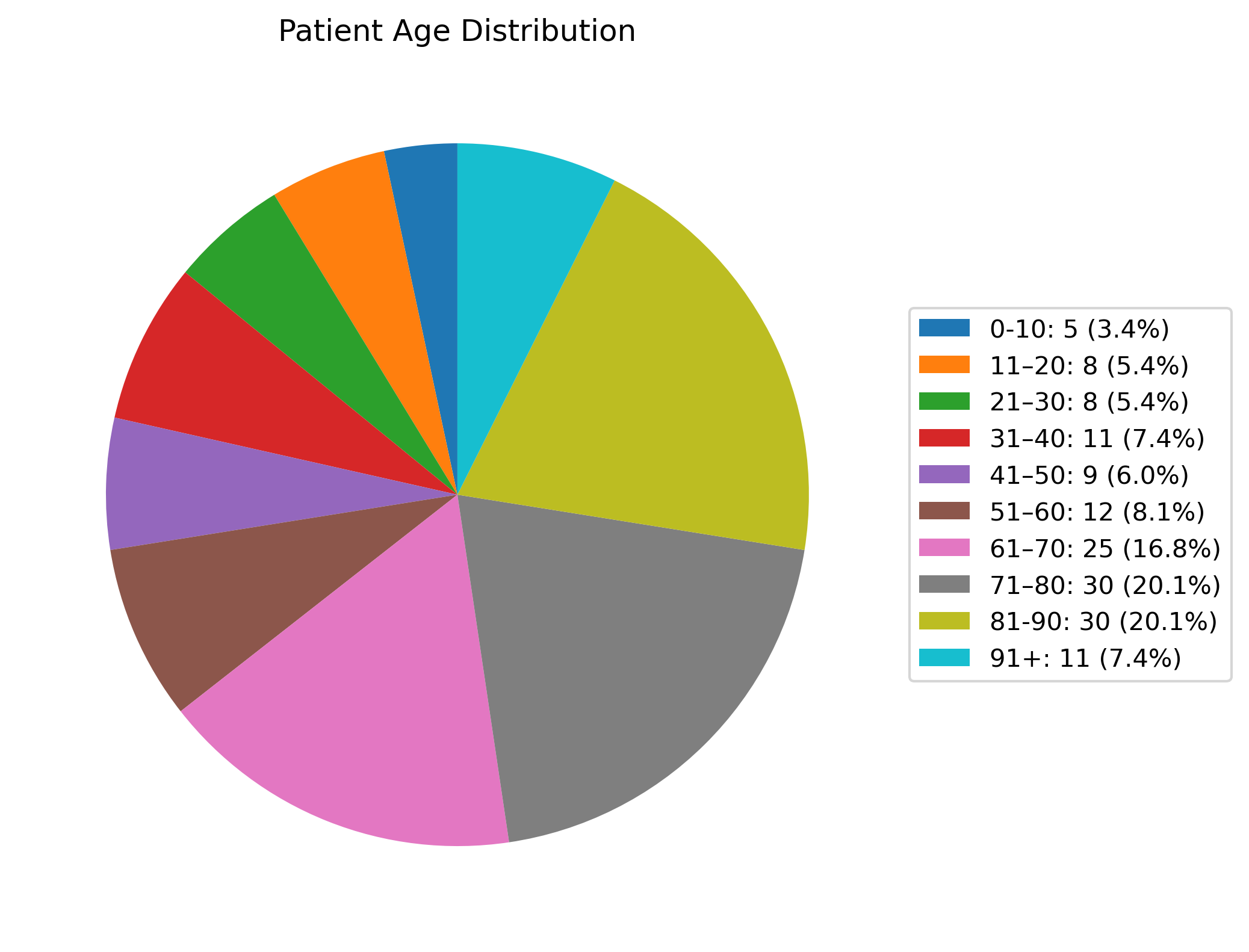}
\label{fig:placeholder}
    \caption{Distribution of patient ages, among known labels, by patient.}
\end{figure}

\subsubsection{Caregiver Category}

Caregiver category is divided in cancer, dementia, both, and miscellaneous, referring to if the caregiver is caring for a care recipient with a diagnosis of the given type.
\begin{figure}[h!]
    \centering
\includegraphics[width=\columnwidth, trim={0cm 0cm 0cm 0cm}, clip]{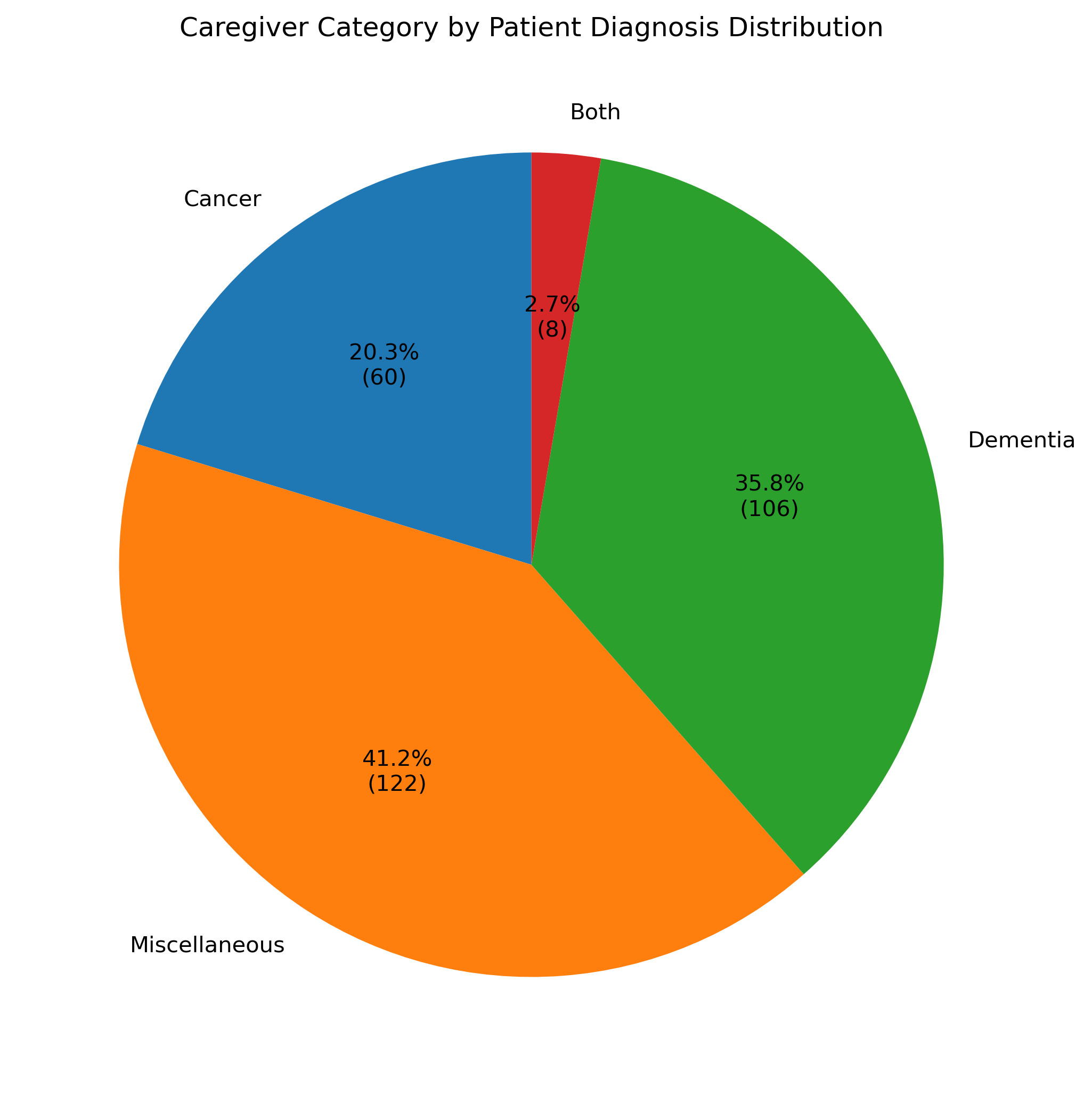}
\label{fig:placeholder}
    \caption{Distribution caregiver categories based on patient diagnosis, among known labels.}
\end{figure}

\subsection{Proportion of Posts with Demographic Attributes Known versus Unknown}

We found the proportion of posts where a given demographic attribute able to be extracted via applying GPT-4o. If an attribute is able to be found, that attribute is considered "known" for that post. 

\begin{table}[htbp]
\centering
\footnotesize
\begin{tabular}{l|cc}
\toprule
Label & Known (\%) & Unknown (\%)\\
\midrule
Caregiver Age        & 25.3 & 74.7 \\
Caregiving Duration           & 43.9 & 56.1 \\
Caregiver Gender     & 25.3 & 74.7\\
\makecell[l]{Caregiver to Patient\\ Relationship}         & 95.8 & 4.2 \\
Patient Age & 32.7 & 67.3 \\
\makecell[l]{Caregiver Category\\ by Patient Diagnosis}  & 78.6 & 21.4\\
\midrule
Overall & 50.267 & 49.733 \\
\bottomrule
\end{tabular}
\caption{The proportion of posts in the caregiver dataset with the demographic attribute known versus unknown.}
\label{tab:annotation-accuracy}
\end{table}

\end{document}